\pdfoutput=1

\documentclass[11pt]{article}

\usepackage[final]{acl}  
\usepackage{amsmath}
\usepackage{cleveref}
\usepackage{amssymb}
\usepackage{amsthm}
\usepackage{bbm}
\usepackage{booktabs}
\usepackage[inline]{enumitem}
\usepackage{scalerel}
\usepackage{subcaption}
\usepackage{tabularx}
\usepackage{tikz}
\usepackage{url}
\usepackage{xurl}
\usepackage{graphicx}
\usepackage{svg}
\usepackage{listings}
\usepackage{xcolor}  
\usepackage{tcolorbox}

\usepackage{algpseudocode}
\usepackage{algorithm}
\usepackage{colortbl} 
\usepackage{times}
\usepackage{latexsym}

\usepackage[T1]{fontenc}
\usepackage{pifont}
\newcommand{\cmark}{\textcolor{green}{\checkmark}} 
\newcommand{\xmark}{\textcolor{red}{\times}}    
\usepackage[utf8]{inputenc}

\usepackage{microtype}

\usepackage{inconsolata}

\usepackage{graphicx}
\usepackage{verbatim}

\usepackage{multirow}

\usepackage{amssymb}
\usepackage{float}

\definecolor{ggreen}{RGB}{52,168,83}

\newcommand{\inc}[1]{{\color{ggreen}(+#1)}}

\title{\textsc{MatViX}: Multimodal Information Extraction from Visually Rich Articles}

\author{\textbf{Ghazal Khalighinejad}$^1$, \textbf{Sharon Scott}$^2$, \textbf{Ollie Liu}$^3$, \textbf{Kelly Anderson}$^2$, \textbf{Rickard Stureborg}$^1$, \\ \textbf{Aman Tyagi}$^2$, \textbf{Bhuwan Dhingra}$^1$ \\
        $^1$\text{Duke University} \\ $^2$\text{The Procter \& Gamble Company} \\
        $^3$\text{University of Southern California}}

\begin{document}
\maketitle
\begin{abstract}
Multimodal information extraction (MIE) is crucial for scientific literature, where valuable data is often spread across text, figures, and tables. In materials science, extracting structured information from research articles can accelerate the discovery of new materials. However, the multimodal nature and complex interconnections of scientific content present challenges for traditional text-based methods. We introduce \textsc{MatViX}, a benchmark consisting of $324$ full-length research articles and $1,688$ complex structured JSON files, carefully curated by domain experts. These JSON files are extracted from text, tables, and figures in full-length documents, providing a comprehensive challenge for MIE.
We introduce an evaluation method to assess the accuracy of curve similarity and the alignment of hierarchical structures. Additionally, we benchmark vision-language models (VLMs) in a zero-shot manner, capable of processing long contexts and multimodal inputs, and show that using a specialized model (DePlot) can improve performance in extracting curves. Our results demonstrate significant room for improvement in current models. Our dataset and evaluation code are available\footnote{\url{https://matvix-bench.github.io/}}.
\end{abstract}

\section{Introduction}

Multimodal information extraction (MIE) has become a key research focus, aiming to extract structured information from both text and visual content~\citep{liu-etal-2019-graph,dong-etal-2020-multi-modal,oka2021machine,sun2024umieunifiedmultimodalinformation}. This is particularly important in scientific literature, where valuable details are often spread across text, figures, and tables. The complex nature of scientific content, combined with the need to combine information from multiple sources, presents substantial challenges for traditional text-based extraction methods.

\begin{figure}[ht]

    \centering
    \hspace*{-3.5mm}                     \includegraphics[scale=0.2]{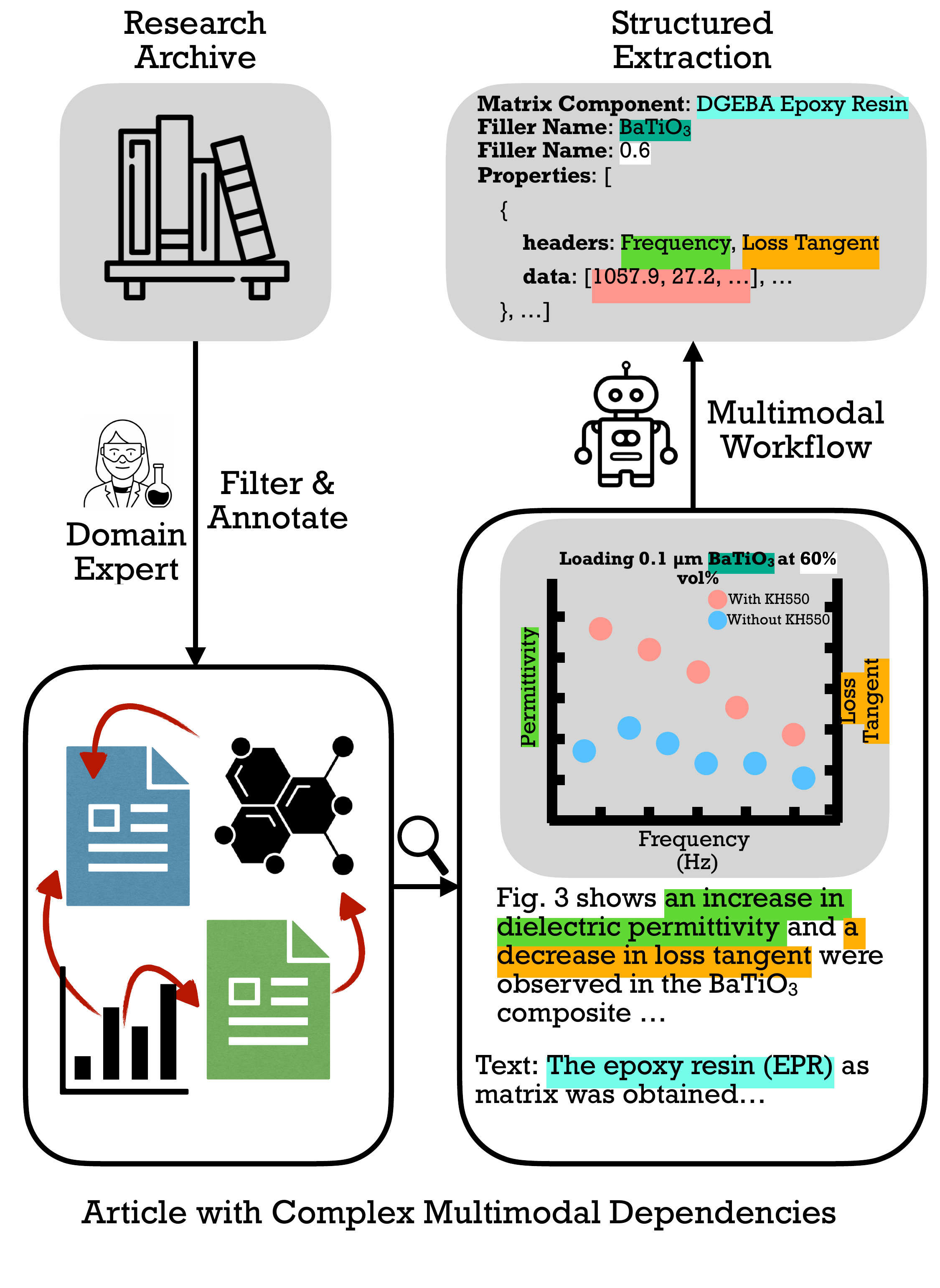}
    \caption{Example of an article with interconnected data between text and figures, with a JSON structure capturing sample properties and composition details.}
    \label{fig:enter-label}
\end{figure}

In materials science, MIE is crucial as research articles contain valuable data that can accelerate the discovery of new materials. Tools like GNoME~\citep{Merchant2023} show how extracting structured databases from these publications can improve discovery efficiency. However, this process is complicated by the multimodal nature of scientific articles and the complex connections between data points. \textbf{Figures are particularly critical, as they often contain essential information about material properties not present in the text}, making their accurate extraction vital for comprehensive information retrieval~\citep{Polak2024}.

Recent methods like DePlot~\citep{liu-etal-2023-DePlot} tackle visual plot reasoning by converting plot images into linearized tables to enable pretrained LLMs to reason over visual data with minimal training. This approach allows LLMs to leverage few-shot reasoning for tasks like chart QA~\citep{masry-etal-2022-chartqa}. However, DePlot's focus on simple plot-to-table conversion makes it less suitable for the complex MIE needed to handle interconnected data spanning long contexts, including text, tables, and multiple figures within full-length scientific documents.

To address this gap, we introduce a novel benchmark, \textsc{MatViX}, focused on obtaining structured information from materials science articles within the domains of polymer nanocomposites and polymer biodegradation. Existing datasets such as FUNSD~\citep{jaume2019funsd}, CORD~\citep{kim2021donut}, and Kleister~\citep{Stanis_awek_2021} have made significant contributions to document analysis and information extraction, especially in handling complex layouts and long documents. However, \textbf{they do not address the complexities of $N$-ary relation extraction}, focusing instead on simpler tasks like named entity recognition (NER), which typically involves identifying predefined entities without capturing intricate relationships between them. Furthermore, these datasets do not consider scientific documents, which often contain specialized language and complex figures. Previous work~\citep{dagdelen2024structured, cheung-etal-2024-polyie} in materials science has explored $N$-ary relation extraction, but primarily from text-only abstracts or short texts. PNCExtract~\citep{khalighinejad-etal-2024-extracting} represents a step forward by considering full-length articles, yet it remains limited to text-based content. In contrast, \textbf{\textsc{MatViX} considers all elements in long scientific documents, including text, tables, and figures}, providing a comprehensive challenge for MIE research (see Table~\ref{tab:dataset_comparison} for a comparison).

Previous work using pretrained models like LayoutLM~\citep{Xu_2020}, LayoutLMv2~\citep{xu2022layoutlmv2}, and domain-specific models such as MatSciBERT~\citep{Gupta2022} has advanced the extraction of structured data from visually rich documents. However, these models are not suitable for our task, as extracting complex $N$-ary relations from long documents exceeds the token limit of $512$ tokens in BERT and LayoutLM models. Additionally, our task involves generating hierarchical JSON files, which is more complex than simple entity recognition or relation extraction and requires a sequence-to-sequence approach. Splitting the documents to fit these models is not feasible, as it disrupts the extraction of interconnected relations across the entire document.

\textbf{\textsc{MatViX} addresses the challenge of long document processing} by utilizing VLMs in a zero-shot manner. These models can process long input contexts~\citep{openai2024gpt4} without sacrificing accuracy and have demonstrated strong performance in tasks requiring both textual and visual reasoning. Given their capabilities, one might wonder if we can simply input a scientific paper, complete with all its images, into a VLM to obtain structured data. In this paper, we benchmark these models and compare them against simpler baselines, showing that there is still substantial room for improvement.

\textbf{Our task involves extracting complex hierarchical structures from scientific documents}, where traditional evaluation metrics fall short. Our method first aligns compositions, which serve as the identity of each sample, and then evaluates the properties represented by curves. To measure the closeness between predicted and ground truth curves, we use the Fréchet distance, which captures how similar the overall trends are. 

Our results show that while VLMs show potential, significant improvements are needed. We also demonstrate that combining the best-performing VLMs with the specialized DePlot model enhances information extraction from figures.

\begin{table*}
\small
\centering
\begin{tabular}{lccccc}
\toprule
\textbf{Dataset} & \textbf{Complex Layout} & \textbf{Long Doc} & \textbf{\textit{N}-ary RE} & \textbf{Scientific} & \textbf{Multimodal} \\
\midrule
FUNSD~\citep{jaume2019funsd} & $\cmark$ &$\xmark$ & $\xmark$ & $\xmark$ & $\xmark$ \\
CORD~\citep{kim2021donut} & $\cmark$ &$\xmark$ & $\xmark$ & $\xmark$ & $\xmark$ \\
Kleister~\citep{Stanis_awek_2021} & $\cmark$ &$\cmark$ & $\xmark$ & $\xmark$ & $\xmark$ \\
PolyIE~\citep{cheung-etal-2024-polyie}  & $\xmark$ & $\xmark$ & $\cmark$ & $\cmark$ & $\xmark$ \\
PNCExtract~\citep{khalighinejad-etal-2024-extracting}  & $\xmark$ & $\xmark$ & $\cmark$ & $\cmark$ & $\xmark$ \\
Ours & $\cmark$ & $\cmark$ & $\cmark$ & $\cmark$ & $\cmark$ \\
\bottomrule
\end{tabular}
\caption{Comparison of Dataset Characteristics. Ours requires \textit{N}-ary relation extraction and is the only dataset that requires extraction from the scientific domain and reasoning over plots.}
\label{tab:dataset_comparison}
\end{table*}

\section{The \textsc{MatViX} Benchmark}

In this section, we first describe our dataset, including the problem definition and dataset preparation, and then explain our evaluation method for the task. \textsc{MatViX} focuses on two critical domains: Polymer Nanocomposites (PNC) and Polymer Biodegradation (PBD). Structured data in these fields is crucial for accelerating research and discovery, as it allows scientists to efficiently analyze relationships between material compositions and properties, which are often spread across text, figures, and tables within research articles. The emphasis on PNC and PBD reflects their significant representation within the field of materials science. The focus on PNC and PBD is justified by their significant presence in the field; a Google Scholar search yields approximately $95,600$ results for ``materials science'', $32,100$ for ``polymer nanocomposites'', and $17,200$ for ``polymer biodegradation''.

Each PNC and PBD sample in the dataset is represented as a structured JSON object that captures both the chemical composition and associated property data. A sample refers to a specific instance of a material with a defined chemical composition and measured properties. This structured format provides detailed information about the materials' compositions and includes the numerical property data needed for training machine learning models. These numerical data points are particularly important for developing models that can predict the relationship between material composition and performance. By leveraging these structured data representations, researchers can conduct large-scale analysis and modeling to advance material discovery and optimization~\citep{Ward2016}.

\subsection{Problem Definition}
Let $\mathcal{D} = {D_1, D_2, \ldots, D_N}$ denote our dataset, which consists of $N$ total articles, where $N=324$. Among these articles, $231$ are from the PNC domain and $93$ are from the PBD domain. 
For each article $D_i \in \mathcal{D}$, there is an associated list of samples $\mathcal{S}_i$, comprising various PNC or PBD samples. Formally, $\mathcal{S}_i$ is defined as:
\begin{equation}
\mathcal{S}_i = {s_{i1}, s_{i2}, \ldots, s_{in_i}},
\end{equation}
where $s_{ij}$ represents the $j$-th sample (either PNC or PBD) in the sample list of the $i$-th article, and $n_i$ denotes the total number of samples in $\mathcal{S}_i$. Each sample $s_{ij}$ is a JSON object. The structure of the JSON object of interest is provided in Appendix~\ref{app:json_formats}.

The goal is to extract the relevant information from each article $D_i$ to populate the corresponding sample list $\mathcal{S}_i$. This involves identifying and extracting the values for each of the entries in the JSON object for every PNC or PBD sample mentioned in the article.

\begin{table}[h]
\small
\centering
\begin{tabular}{lcc}
\toprule
\textbf{Statistic} & $\mathcal{D}_{PNC}$ & $\mathcal{D}_{PBD}$ \\ \midrule
Total Papers & 231 & 93 \\ 
Total Samples & 1396 & 292 \\ 
Avg. Samples per Paper  & 6 & 3 \\ 
Avg. Tokens per Paper & 8905 & 8456 \\ 
\bottomrule
\end{tabular}
\caption{Data Statistics for $\mathcal{D}_{PNC}$ and $\mathcal{D}_{PBD}$}
\label{tab:data_stats}
\end{table}

\subsection{Polymer Nanocomposites (PNC)}
\subsubsection{Overview}
Our PNC dataset, derived from the Nanomine data repository~\citep{10.1063/1.5046839}, extends PNCExtract~\citep{khalighinejad-etal-2024-extracting} by including both compositions and properties of PNC samples. 

Each PNC sample $s_{ij}$ is represented as a structured JSON object comprising two main sections: composition and properties. The composition section specifies the matrix and filler materials along with their attributes, while the properties section contains specific characteristics of the sample, including their names, measurement conditions, and corresponding data points (see Appendix~\ref{app:json_formats} for the JSON format).

We focus on six key properties frequently studied in the dataset: Thermal, Electrical, Mechanical, Viscoelastic, Volumetric, and Rheological. These properties are prioritized because they are not only the most commonly reported in research papers but also critical in determining the performance of polymer nanocomposites. Each property includes numerical data collected under various experimental conditions, specified in the JSON headers, with the actual data points listed. Examples of property representations and their associated plots are provided in Appendix~\ref{app:example_pnc_properties}.

An analysis of the Nanomine data repository reveals the distribution of these properties across $4,186$ samples:  Thermal (26.4\%), Electrical (29.6\%), Mechanical (14.1\%), Viscoelastic (21.0\%), Volumetric (3.3\%), Rheological (5.4\%), and Others (0.1\%). Therefore, we concentrate on the six key properties due to their significance in the dataset.

\subsubsection{Data Collection}
Our dataset is based on data from the NanoMine repository~\citep{10.1063/1.5046839}, a comprehensive resource for PNC data structured around an XML-based schema for the representation and sharing of nanocomposite materials information. The original data in NanoMine was collected and stored using Excel templates provided to materials researchers. However, this structure is not consistent and includes a large template with $43$ attributes in the Materials Composition section and over $20$ different properties, all organized in formats that are challenging to process.

To address these inconsistencies, we standardized and cleaned the NanoMine repository data. We categorized the $20$ properties into six main categories—Thermal, Electrical, Mechanical, Viscoelastic, Volumetric, and Rheological—ensuring that the data points within these properties were restructured and aligned accordingly. The process involved organizing the data into a structured JSON format suitable for our analysis and modeling purposes.

This categorization and cleaning effort were validated by the experts in the field, to ensure the accuracy of the structured data.

\subsection{Polymer Biodegradation (PBD)}
\subsubsection{Overview}
Our PBD dataset focuses on extracting information related to the biodegradation of polymers. The dataset was collected by experts in biodegradable polymers, ensuring high accuracy through meticulous data collection and verification. The dataset includes $47$ research papers in the test set and $46$ papers in the validation set, with a total of $159$ samples for testing and $133$ samples for validation.

Each PBD sample $s_{ij}$ is captured as a structured JSON object that details both the composition and biodegradation characteristics of the polymer sample. The structure captures essential information regarding the polymer's type, characteristics, and biodegradation data, including conditions and corresponding measurements. For the detailed structure of the JSON format, please refer to Appendix~\ref{app:json_formats}. The biodegradation results are typically presented in figures, showing plots of conditions versus biodegradation percentage (refer to Appendix~\ref{app:example_pbd_properties}).

\subsubsection{Data Collection}
Two materials science experts curated a collection of $93$ research papers focused on biodegradable materials, selecting high-quality articles from reputable journals. They first identified key compositional attributes consistent across polymer biodegradation samples. One expert extracted these details, while the second verified their accuracy.

After validating compositions, the experts extracted properties, which were often presented in text, tables, or figures. For plots—commonly showing biodegradation rates—the PlotDigitizer\footnote{https://plotdigitizer.com/} tool was used to trace curves and extract (x, y) data points. This process involved calibrating axes and converting visual information into structured JSON files. For a detailed explanation of the data collection and digitization process, see the Appendix~\ref{app:annotation_guidelines}.

\section{Evaluation}

Each paper contains a set of samples, and each sample is characterized by its composition and properties. The composition is represented as a set of strings, while the properties are captured as a list of curves. To evaluate the accuracy of the predicted samples against the ground truth, we follow a two-step process: first, we evaluate the alignment of the compositions within the samples, and then we assess the matching of the properties for the aligned samples. The reason for this approach is that the composition defines what the sample is, providing its identity, while the properties describe the characteristics of that specific sample. Therefore, it is crucial to first match the compositions correctly, ensuring that we are comparing the same types of samples, before evaluating the properties within those matched samples.

The evaluation employs the F1 Score for composition alignment and two specialized metrics, the Curve Similarity Score (CSS) and the Curve Alignment Score (CAS), for property evaluation.

\paragraph{Composition Alignment}
To assess the alignment between predicted and ground truth compositions within each sample, we treat this as a maximum bipartite matching problem. Each composition consists of a set of strings, and we aim to find the best correspondence between the predicted and ground truth compositions. We use the Munkres algorithm (Hungarian algorithm)~\citep{https://doi.org/10.1002/nav.3800020109} to solve this bipartite matching problem, optimizing for the highest possible F1 Score. If there are more ground truth samples for a paper than predicted samples, the unmatched ground truth samples are considered false negatives. Conversely, if there are more predicted samples than ground truth samples, the unmatched predicted samples are considered false positives.

\paragraph{Curve Similarity Score (CSS)}
Once compositions are aligned, we evaluate the properties within these matched samples. Each property is represented by a curve, which is defined as a list of (x, y) points, and the CSS is introduced as a quantitative measure of similarity between the predicted and ground truth properties. These properties are typically plotted as the relationship between a variable and its corresponding response, such as how dielectric permittivity changes with frequency or temperature, which are commonly reported in research papers. The trends captured in these curves are crucial, as they convey important information about the material's behavior. Therefore, accurately extracting and evaluating these curves is essential.

To quantify the similarity, we use the Levenshtein distance to compare the headers (x and y-axis labels) and the Fréchet distance to compare the ground-truth data points of the curves. The Fréchet distance measures the similarity between two curves by finding the smallest of the maximum pairwise distances. To compute this for polygonal curves, the discrete Fréchet distance is used, as shown by \citet{Wien94computingdiscrete}, which involves determining the shortest path through a coupling sequence that matches points between the curves while maintaining their order. 

The CSS, given a predicted curve \( c_p \) and a ground truth curve \( c_t \), is calculated as follows:

\begin{align}
    \begin{split}
        \text{CSS}(c_p, c_t) = & \left(1 - \text{nl}_{\text{lev}}\left(d_{\text{lev}}(h_p, h_t)\right)\right) \\
        & \left(1 - \text{nl}_{\text{frech}}\left(d_{\text{frech}}(c_p, c_t)\right)\right),
    \end{split}
\end{align}

where \( d_{\text{lev}}(h_p, h_t) \) is the Levenshtein distance between the headers of the predicted (\( h_p \)) and ground truth (\( h_t \)) curves, and \( d_{\text{frech}}(c_p, c_t) \) is the Fréchet distance between the predicted and ground truth curves. The normalization functions \( \text{nl}_{\text{lev}} \) and \( \text{nl}_{\text{frech}} \) are defined as follows:

\begin{align*}
    \text{nl}_{\text{lev}}(d_{\text{lev}}) &= \min\left(1, \frac{d_{\text{lev}}(h_p, h_t)}{\max(\text{len}(h_p), \text{len}(h_t))}\right) \\
    \text{nl}_{\text{frech}}(d_{\text{frech}}) &= \min\left(1, \frac{d_{\text{frech}}(c_p, c_t)}{\|c_t\|}\right)
\end{align*}

where \( \text{len}(h) \) represents the length of the header, and \( \|c_t\| \) denotes the norm of the ground truth curve data. 

This approach addresses several limitations inherent in the metric introduced in DePlot, which misses critical information about the alignment of trends. By incorporating the Fréchet distance, the CSS provides a more comprehensive evaluation, capturing both the trend similarities.

\paragraph{Curve Alignment Score (CAS)}
The CAS metric identifies the best match between predicted and ground truth curves when multiple curves are present within a sample. Let \( X \in \mathbb{R}^{N \times M} \) be a binary matrix where \( X_{ij} \) indicates the assignment of the \( i \)-th predicted curve to the \( j \)-th ground truth curve, based on the Munkres algorithm. The CAS is calculated as follows:

\[
\text{CAS} = \frac{1}{\max(N, M)} \sum_{i=1}^{N} \sum_{j=1}^{M} X_{ij} \cdot \text{CSS}(c_{p_i}, c_{t_j}),
\]

where \( \text{CSS}(c_{p_i}, c_{t_j}) \) represents the Curve Similarity Score between the \( i \)-th predicted curve (\( c_{p_i} \)) and the \( j \)-th ground truth curve (\( c_{t_j} \)). 


\subsection{Human Evaluation}
We conduct a human evaluation study to assess the effectiveness of our evaluation metric. A total of $50$ plot-prediction pairs (GPT-4o predictions) are randomly sampled from our dataset, representing a range of different scores.

Three human annotators, each with relevant expertise in the field, evaluate these pairs. For each sample, the annotators are presented with both the ground-truth plot and the model's predicted plot. They are asked to assess the quality of the prediction based on two specific questions: (1) Are the axes labeled correctly? (2) Is the trend of the predicted curve consistent with the ground truth? (See Appendix~\ref{app:human_eval_guideline} for details.)

The human scores for the first question (regarding headers) were averaged and compared to the automated header scores after alignment, calculated as $1 - \text{nl}_{\text{lev}}\left(d_{\text{lev}}(h_p, h_t)\right)$. Similarly, the human scores for the second question (regarding curves) were averaged and compared to the automated curve scores after alignment, calculated as $1 - \text{nl}_{\text{frech}}\left(d_{\text{frech}}(c_p, c_t)\right)$. 

For comparison, we used both Pearson’s $r$ and Spearman’s $\rho$ correlation coefficients. Table~\ref{tab:correlation_results} indicates positive correlations between the human judgments and the automated metric scores.

\begin{table}[ht]
\centering
\small
\begin{tabular}{lcc}
        \toprule
        & \textbf{Coefficient} & \textbf{p-value} \\
        \midrule
        Pearson $r$ (curves) & 0.887 & $4.17 \times 10^{-7}$ \\
        Spearman $\rho$ (curves) & 0.717 & 0.00055 \\
        Pearson $r$ (headers) & 0.930 & $8.54 \times 10^{-9}$ \\
        Spearman $\rho$ (headers) & 0.921 & $2.28 \times 10^{-8}$ \\
        \bottomrule
    \end{tabular}
\caption{Correlation results between model scores and human evaluations for header and curve scores, using Pearson's $r$ and Spearman's $\rho$. Human scores represent the average ratings from three annotators.}
\label{tab:correlation_results}
\end{table}

\section{Benchmarking VLMs}
The objective is to extract structured data from materials science documents. To achieve this, we first convert each PDF document into LaTeX format using Mathpix (https://mathpix.com/). This approach generates a TeX file that shows the structure of the paper, including sections, subsections, and all images.
We then employ Visual Language Models (VLMs) in a zero-shot manner to extract structured JSON data. 

During our preliminary experiments, we observed that providing both the entire LaTeX file and all associated images as input to the VLMs leads to suboptimal results. Additionally, since we must input images one at a time, it becomes costly 
To address this, we devised a multi-step pipeline (see Appendix~\ref{app:nanocomposite_extraction}):

\begin{itemize}
    \item \textbf{Text Information Extraction:} First, we use an LLM to extract structured information from the text in the LaTeX document.
    \item \textbf{Information Expansion:} For each image, we then prompt the VLM to expand the extracted information based on the text and the images. This expansion is handled individually for each image.
    \item \textbf{Information Integration:} Given that multiple images are typically associated with a document, we merge all the expanded information from the different images to create a comprehensive, structured dataset.
\end{itemize}

Formally, the steps can be described as follows:

\begin{align}
    \hat{S}_i^{\text{text}} &= \text{LLM}(D_i^{\text{text}}) \\
    \hat{S}_i^{\text{img}, k} &= \text{VLM}(\hat{S}_i^{\text{text}}, I_k), \quad \forall k \in [K] \\
    \hat{S}_i &= \text{Merge}(\hat{S}_i^{\text{text}}, \{\hat{S}_i^{\text{img}, k}\}_{k=1}^K)
\end{align}

where $D_i^{\text{text}}$ is the textual data in document $i$, $\hat{S}_i^{\text{text}}$ is the predicted sample list derived only from the textual data of document $i$, and $I_k$ is the $k$th image in the document. $\hat{S}_i^{\text{img}, k}$ represents the expanded information obtained by the VLM for the $k$-th image using $\hat{S}_i^{\text{text}}$ as context. Finally, $\hat{S}_i$ merges the textual and image-based information to form a comprehensive structured dataset for document $i$.

\section{Experiments}

In this section, we present the results of modeling with VLMs on \textsc{MatViX}.

\begin{table*}[t!]
\centering
\small
\begin{tabular}{ll|p{10mm} p{10mm} p{10mm}|p{10mm} p{10mm} p{17mm}}
\toprule
\textbf{Model} & \textbf{Config} & \textbf{P} & \textbf{R} & \textbf{F1} & \textbf{Headers} & \textbf{Curves} & \textbf{CAS} \\
\midrule

\rowcolor{lightgray} \multicolumn{8}{c}{\textbf{Polymer Nanocomposite Dataset}} \\
\midrule
\multirow{3}{*}{\textbf{GPT-4o}} 
& Base  & -- & -- & -- & \textbf{16.31} & \textbf{18.54} & \textbf{6.53} \\
& T-Only  & 67.78 & 42.67 & 51.88 & 10.55 & 05.64 & 1.48 \\
& T-Only + DePlot & 62.07 & 39.05 & 47.41 & 10.86 & 04.63 & 1.10 \\
& T+Img  & \textbf{68.33} & \textbf{43.24} & \textbf{52.47} & 13.25 & 12.58 & 2.43 \\
& T+Img + DePlot & 62.74 & 39.67 & 48.11 & 14.21 & 12.54 & 2.31 \\
\midrule

\multirow{3}{*}{\textbf{GPT-4-Turbo}} 
& Base  & -- & -- & -- & \textbf{16.97} & 17.27 & \textbf{6.51} \\
& T-Only  & 70.35 & 46.36 & 55.70 & 12.88 & 09.14 & 2.46 \\
& T-Only + DePlot & \textbf{73.08} & \textbf{47.98} & 55.70 & 14.22 & 15.81 & 4.94 \inc{{2.48}} \\
& T+Img  & 68.34 & 43.14 & 52.41 & 14.92 & 10.67 & 1.87 \\
& T+Img + DePlot & 72.25 & 47.54 & \textbf{57.11} & 16.45 & \textbf{17.28} & 4.46 \inc{2.59}\\
\midrule

\multirow{3}{*}{\textbf{Claude 3.5}} 
& Base  & -- & -- & -- & 15.08 & 14.56 & \textbf{5.19} \\
& T-Only  & \textbf{55.55} & \textbf{35.95} & \textbf{43.33} & 13.74 & 10.99 & 3.45 \\
& T-Only + DePlot & 50.40 & 33.13 & 39.75 & 14.93 & 15.16 & 4.76  \inc{1.31}\\
& T+Img  & 52.45 & 33.16 & 40.21 & 15.50 & 18.36 & 4.50 \\
& T+Img + DePlot & 48.14 & 30.91 & 37.33 & \textbf{16.04} & \textbf{18.51} & 4.93 \inc{0.43} \\
\midrule

\multirow{3}{*}{\textbf{Claude 3}} 
& Base  & -- & -- & -- & 17.00 & 15.40 & 5.74 \\
& T-Only  & \textbf{51.93} & \textbf{33.03} & \textbf{40.07} & 11.06 & 04.88 & 1.54 \\
& T-Only + DePlot & 50.00 & 31.40 & 38.19 & \cellcolor{yellow!40}\textbf{18.93} & 11.02 & 3.37 \inc{1.83} \\
& T+Img  & 51.46 & 31.62 & 38.74 & 12.80 & 16.88 & 3.79 \\
& T+Img + DePlot & 44.72 & 27.94 & 34.09 & 16.60 & \textbf{23.21} & \textbf{6.55} \inc{2.76} \\
\midrule

\multirow{3}{*}{\textbf{Gemini 1.5}} 
& Base  & -- & -- & -- & 17.19 & 17.13 & \cellcolor{yellow!40}\textbf{6.73} \\
& T-Only  & \cellcolor{yellow!40}\textbf{72.28} & \cellcolor{yellow!40}\textbf{47.69} & \cellcolor{yellow!40}\textbf{57.37} & \textbf{18.29} & 07.07 & 2.70 \\
& T-Only + DePlot & 68.11 & 44.84 & 54.00 & 17.33 & 14.28 & 4.34 \inc{1.64} \\
& T+Img  & 71.17 & 46.76 & 56.32 & 14.96 & 19.08 & 3.98 \\
& T+Img + DePlot & 66.96 & 44.10 & 53.10 & 16.55 & \cellcolor{yellow!40}\textbf{24.10} & 5.25 \inc{1.27} \\
\midrule

\rowcolor{lightgray} \multicolumn{8}{c}{\textbf{Polymer Biodegradation Dataset}} \\
\midrule
\multirow{3}{*}{\textbf{GPT-4o}} 
& Base  & -- & -- & -- & \textbf{93.39} & 15.66 & 14.57 \\
& T-Only  & 35.53 & 18.80 & 23.60 & 55.37 & \textbf{31.07} & \textbf{21.78} \\
& T-Only + DePlot & 40.86 & 20.86 & 26.48 & 49.33& 30.83 & 19.53 \\
& T+Img  & 35.99 & 18.90 & 23.76 & 43.70 & 28.60 & 15.18 \\
& T+Img + DePlot & \textbf{41.81} & \cellcolor{blue!30}\textbf{23.00} & \cellcolor{blue!30}\textbf{28.40} & 37.75 & 26.94 & 11.46 \\
\midrule

\multirow{3}{*}{\textbf{GPT-4-Turbo}} 
& Base  & -- & -- & -- & \textbf{94.84} & 15.56 & 15.04 \\
& T-Only  & 26.58 & 16.37 & 19.37 & 46.49 & 23.78 & 13.43 \\
& T-Only + DePlot & 26.72 & 16.21 & 19.41 & 49.38 & \textbf{35.37} & \textbf{16.24} \inc{2.81}\\
& T+Img  & \textbf{31.50} & \textbf{20.61} & \textbf{23.98} & 39.45 & 26.76 & 13.21 \\
& T+Img + DePlot & 27.77 & 16.66 & 20.05 & 40.33 & 30.49 & 11.54 \\
\midrule

\multirow{3}{*}{\textbf{Claude 3.5}} 
& Base  & -- & -- & -- & \textbf{94.03} & 16.57 & 16.09 \\
& T-Only  & \textbf{26.12} & \textbf{13.46} & \textbf{16.70} & 47.82 & 33.11 & 20.07 \\
& T-Only + DePlot & 24.08 & 13.07 & 15.82 &47.32 & \cellcolor{blue!30}\textbf{38.92} & \cellcolor{blue!30}\textbf{22.52} \inc{2.45}\\
& T+Img  & 24.25 & 12.27 & 15.27 & 30.85 & 21.15 & 11.21 \\
& T+Img + DePlot & 22.87 & 12.50 & 15.03 & 20.94 & 22.30 & 09.23 \\
\midrule

\multirow{3}{*}{\textbf{Claude 3}} 
& Base  & -- & -- & -- & \cellcolor{blue!30}\textbf{95.56} & 16.81 & \textbf{16.35} \\
& T-Only  & 41.66 & 20.25 & 26.01 & 24.90 & \textbf{20.21} & 06.78 \\
& T-Only + DePlot & 33.69 & 19.28 & 23.69 & 17.58 & 18.03 & 03.52 \\
& T+Img  & \cellcolor{blue!30}\textbf{44.71} & \textbf{21.12} & \textbf{27.29} & 13.98 & 18.09 & 03.30 \\
& T+Img + DePlot & 38.51 & 21.21 & 26.19 & 18.44 & 12.22 & 01.56 \\
\midrule

\multirow{3}{*}{\textbf{Gemini 1.5}} 
& Base  & -- & -- & -- & \textbf{93.20} & 17.61 & \textbf{16.69} \\
& T-Only  & \textbf{23.08} & \textbf{16.55} & \textbf{18.52} & 52.36 & \textbf{25.63} & 16.02 \\
& T-Only + DePlot & 17.87 & 10.84 & 12.85 & 44.92 & 25.21 & 15.55 \\
& T+Img  & 18.98 & 12.70 & 14.43 & 18.29 & 05.65 & 02.97 \\
& T+Img + DePlot & 18.51 & 12.15 & 14.06 & 16.03 & 08.54 & 03.37 \inc{0.4} \\
\bottomrule
\end{tabular}
\caption{Evaluation results for predicting \textbf{Compositions} (P: Precision, R: Recall, F1: F1-Score) and \textbf{Properties} (Headers, Curves, CAS) under different configurations (Base: Baseline, T-Only: Text Only, T+Img: Text + Image). The highlighted values indicate the highest scores among the models. {\color{ggreen}(green)} indicates the increase in performance when using DePlot compared to its non-DePlot counterpart in the same configuration.}
\label{tab:main_results_no_deplot}
\end{table*}

\subsection{Models and Setup}

We use GPT-4-Turbo, GPT-4o, Claude-3-Haiku, Claude-3.5-Sonnet, and Gemini-Pro-1.5 in our experiments~\citep{openai2024gpt4, Claude3, reid2024gemini} which are instruction-tuned models and are prompted in a zero-shot manner. We also conducted preliminary experiments with the open-sourced Vicuna-7b-v1.5-16k~\citep{vicuna2023} model, but it failed to capture any meaningful structure. We evaluate these models against a custom baseline approach called Majority Vote. The baseline method selects the most common header and curve predictions from the validation set.

\subsection{Baseline for Headers, Curves, and CAS}

First, for the curves associated with each property (note that there are six properties in the PNC dataset), we calculate the average Fréchet Distance among all curves in the validation set. The curve that is the closest to all others (i.e., has the smallest cumulative Fréchet Distance) is selected as the baseline curve for that property. For the headers, we examine the validation set to determine the most common x-header and y-header for each property. These most frequent headers serve as the baseline headers.

Next, we consider the predictions from the LLM ($\hat{S}_i^{\text{text}}$). We then add the baseline properties to each predicted composition. Specifically, for PNC, we calculate the average occurrence of each of the six properties per sample and include that many for each property. For PBD, we expand by adding one baseline property directly, as there is only one property.

\subsection{Results} 
Table~\ref{tab:main_results_no_deplot} presents model performance in composition extraction, curve and header extraction, and curve alignment.

\paragraph{Challenges in Curve Alignment.} 
Across both PNC and PBD datasets, models demonstrate stronger performance in composition extraction, with Gemini 1.5 achieving the highest F1-Scores on the PNC dataset and GPT-4o leading on PBD. Curve extraction is more challenging, particularly in the PNC domain; the best model achieves only $6.55$. This lower performance reflects the complexity of curve extraction, as it requires interpreting data from tables and figures, extracting curves accurately, and then aligning them with the appropriate composition. This process is inherently more complex than composition extraction, which requires fewer reasoning steps and draws from more straightforward data sources.

\paragraph{Baseline Often Outperforms in CAS.} The baseline configuration sometimes provides the best results for CAS. However, there are instances where models outperform the baseline, such as Claude 3.5 on the PBD dataset. Interestingly, most models outperform the baseline in curve extraction but struggle in header extraction, which reveals a gap in their ability to fully integrate and interpret all data components.

\paragraph{T+Img Configuration Does Not Always Enhance Curve Extraction.} Surprisingly, incorporating both text and images (T+Img) does not consistently lead to better performance in curve extraction. While some information is only present in the images and not in the text, current VLMs seem more influenced by the noise from the images than by the useful data they contain. As a result, the T-Only configuration is often more effective, as it relies on focused textual information without the interference introduced by noisy visual inputs. Note that, in the T-Only case, since the entire LaTeX file is provided as input to the LLM, the tables are included, and in many cases, the important results from the figures are mentioned in the tables.

\subsubsection{Specialized Tools for Chart-to-Text Extraction.} While specialized tools for curve extraction exist, they are insufficient when information is interconnected with text, tables, and figures. Tools like PlotDigitizer can manually extract data points but lack automation, requiring annotators to calibrate axes and mark values manually. To our knowledge, no fully automated solutions are available.

We further test whether integrating DePlot with models can enhance results. Our approach involves first passing all images through DePlot to obtain its output. We then replace the figure section of the LaTeX file with DePlot's output, which is formatted as a linearized table. This modified LaTeX file is fed into an LLM for evaluation. For the T+Img configuration, we provide the LLM's prediction along with the original images to assess the combined performance. Table~\ref{tab:main_results_no_deplot} shows that using DePlot can improve the results of the best-performing configurations. However, DePlot sometimes hurts performance, particularly on the PBD dataset. We hypothesize that this is because most curve-related information in the PBD dataset is embedded in the main text or tables rather than images, making DePlot's output less relevant and potentially adding noise.

\section{Related Work}

A plethora of prior works have devoted to unimodal information extraction (IE) with LLMs. See \citep[\textit{inter alia}]{tchoua2019creating, oka2021machine, xie2023large, shetty2023general, dagdelen2024structured} for an overview of their applications in scientific texts. In contrast, there lacks specialized IE systems that jointly operate on scientific documents that contain texts, tables, and images \citep{dong-etal-2020-multi-modal,gupta2022discomat,sun2024umieunifiedmultimodalinformation}. 

General-purpose foundation models \citep{openai2024gpt4, reid2024gemini, Claude3} are appealing alternatives for such tasks; yet directly applying these models often yields subpar performance due to the complexity of document structures \cite{khalighinejad-etal-2024-extracting}, their inability to reason over long contexts and/or multiple images \cite{reid2024gemini}, and performance differences across modalities \citep{li2024mmsci, fu2024isobench}.

Several recent works have endeavoured to adapt pre-trained foundation models for material science \citep{Gupta2022,gupta2022discomat,song2023honeybee}, but extending this fine-tuning approach to images has been challenging. This is attributed to subpar performance of open-weight VLMs \cite{yue2023mmmu} -- noted for their lack of faithfulness \citep{fu2024tldr} and compositionality \citep{kong2023interpretable} compared to API-access models -- as well as a lack of high-quality multimodal datasets in the material science domain \cite{miret2024llms}. \textsc{MatViX} aims to bridge this gap by contributing an expert-annotated, multimodal dataset over full-length scientific documents, and a workflow that achieves nontrivial performance according to a curated evaluation suite across different modalities.

In this respect, \textsc{MatViX} connects more broadly to a growing body of works that evaluate LLMs as agentic systems \citep{liu2023agentbench, mialon2023gaia,koh-etal-2024-visualwebarena,liu2024dellma,xie2024osworld}. Compared to knowledge-intensive benchmarks, they arguably evaluate model capabilities more akin to daily workflows, and are robust against data contamination. \textsc{MatViX} subsidizes this research with a dataset driven by scientific use cases, and offers a suite of \textit{partial} evaluation metrics that enable users to identify areas of improvements compared to binary success metrics.

\section{Conclusion and Future Work}

We introduce \textsc{MatViX}, an expert-annotated, multimodal information extraction benchmark developed from scholarly articles. These articles receive a minimal amount of pre-processing; they are thus endowed with diverse textual, tabular, and visual structures, all of which contain information important for scientific applications. A general workflow is proposed, and is evaluated against a suite of automatic evaluation metrics that ensure the accuracies of extracted data across all modalities. Results validate the performance of our workflow, and our automatic metrics agree with human evaluations.

There are many avenues for future work. One such example is exploring an agentic framework where the model utilizes various smaller models or tools to assist with the extraction task. For instance, as shown in Table~\ref{tab:main_results_no_deplot}, DePlot is helpful for image-to-table extraction. Additionally, we hypothesize that a BERT model specifically trained for NER and RE may achieve higher recall than generative LLMs. Therefore, integrating these components into an agentic framework could be a promising next step. Another direction is to validate the usefulness of extracted information for domain scientists. In materials science, much of the extracted data is used to train downstream machine learning models~\citep{xu2023small}. To assess the effectiveness of our system, we can compare the performance of models trained on the extracted data with those trained on ground truth data.

\section{Limitations}
The \textsc{MatViX} benchmark provides valuable insights into multimodal information extraction in the PNC and PBD domains. However, there are several limitations to consider.

First, our benchmark is limited to these two specific domains within materials science. While this focus is important for advancing research in these fields, the findings may not generalize well to other scientific disciplines. Future work should explore expanding the dataset to include additional areas of science.

Additionally, we only considered a zero-shot approach in this paper. While this is effective for evaluating the generalization capabilities of VLMs, fine-tuning these models on domain-specific data could further improve their performance, though this was outside the scope of our current study.

Finally, our evaluation metrics, particularly for curve extraction, do not take into account the units of measurement, which can be critical for scientific analysis. While the Fréchet distance helps measure trend similarity, the absence of unit considerations limits the metric's ability to fully assess the accuracy of the extracted data. Future work should explore more domain-specific metrics that account for both trends and units to provide a deeper understanding of model performance.

\section{Ethics Statement}
We do not believe there are significant ethical issues associated with this research.

\section{Acknowledgement}
This research was supported by a gift from Procter \& Gamble.

\appendix
\label{sec:appendix}

\section{Terms of Use}
We used OpenAI Models, Claude, Gemini, and the NanoMine data repository in accordance with their licenses and terms of use.

\section{Computational Experiments Details}
\paragraph{Hyperparameter Settings}
The models used in our experiments, OpenAI Models, Claude, and Gemini, have been evaluated for their performance in multimodal information extraction tasks within the \textsc{MatViX} benchmark while the temperature parameter is set to zero to ensure consistent evaluation.

\section{JSON Formats}
\label{app:json_formats}

\paragraph{Polymer Nanocomposites}
\begin{verbatim}
{
    "Matrix Component": "",
    "Matrix Abbreviation": "",
    "Filler Chemical Name": "",
    "Filler Abbreviation": "",
    "Filler PST": "",
    "Filler Mass": "",
    "Filler Volume": "",
    "Properties": [
        {
            "data": [
                ["", ""],
                ["", ""]
            ],
            "headers": [
                "x-label",
                "y-label"
            ],
            "property name": ""
        },
    ]
}
\end{verbatim}

\paragraph{Polymer Biodegradation}
\begin{verbatim}
{
    "Polymer Type": "",
    "Substitution Type": "",
    "Degree of Substitution": "",
    "Comonomer Type": "",
    "Degree of Hydrolysis": "",
    "Molecular Weight": "",
    "Molecular Weight Unit": "",
    "Biodegradation Test Type": "",
    "Biodegradation": {
        "header": [
            "x-label",
            "y-label"
        ],
        "data": [
            [x1, y1],
            [x2, y2],
            ...
        ]
    }
}
\end{verbatim}

\begin{figure*}
\section{Properties Examples}
\subsection{PNC}
    \centering
    \begin{subfigure}[b]{0.45\textwidth}
        \centering
        \includegraphics[width=\textwidth]{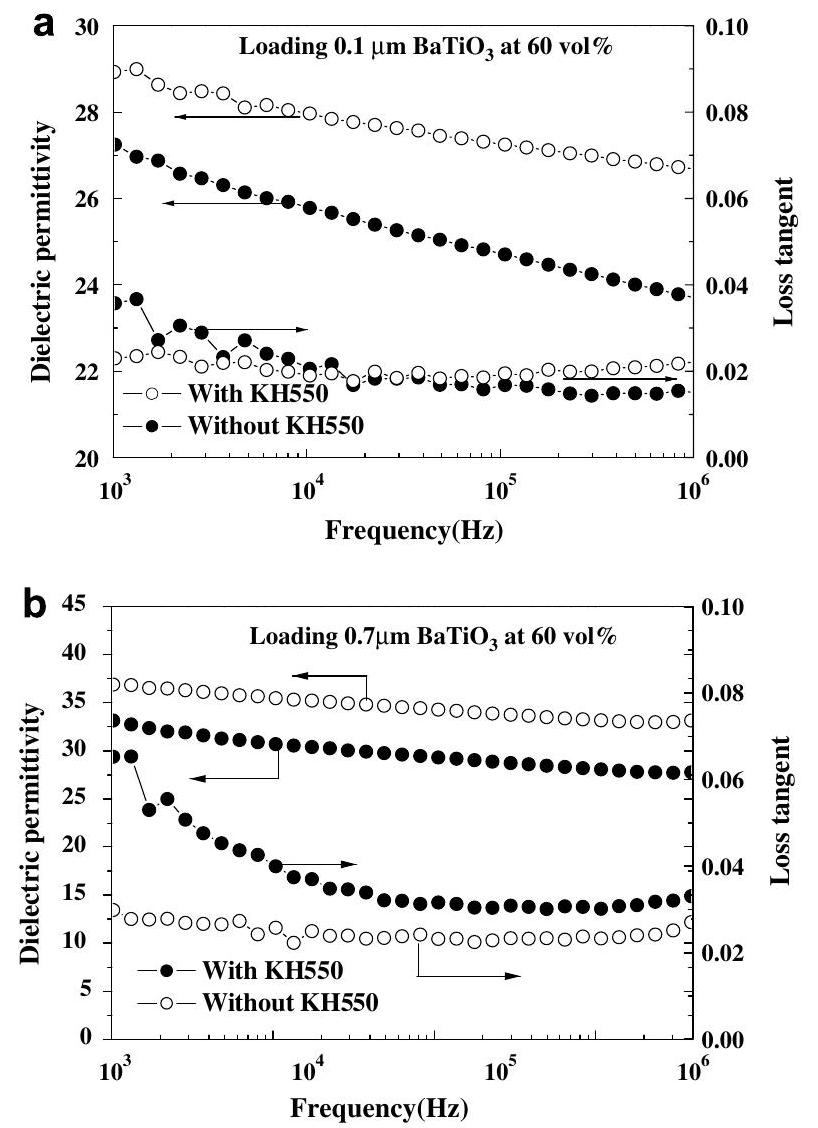}
    \end{subfigure}
    \hfill
    \begin{subfigure}[b]{0.45\textwidth}
        \centering
        \includegraphics[width=\textwidth]{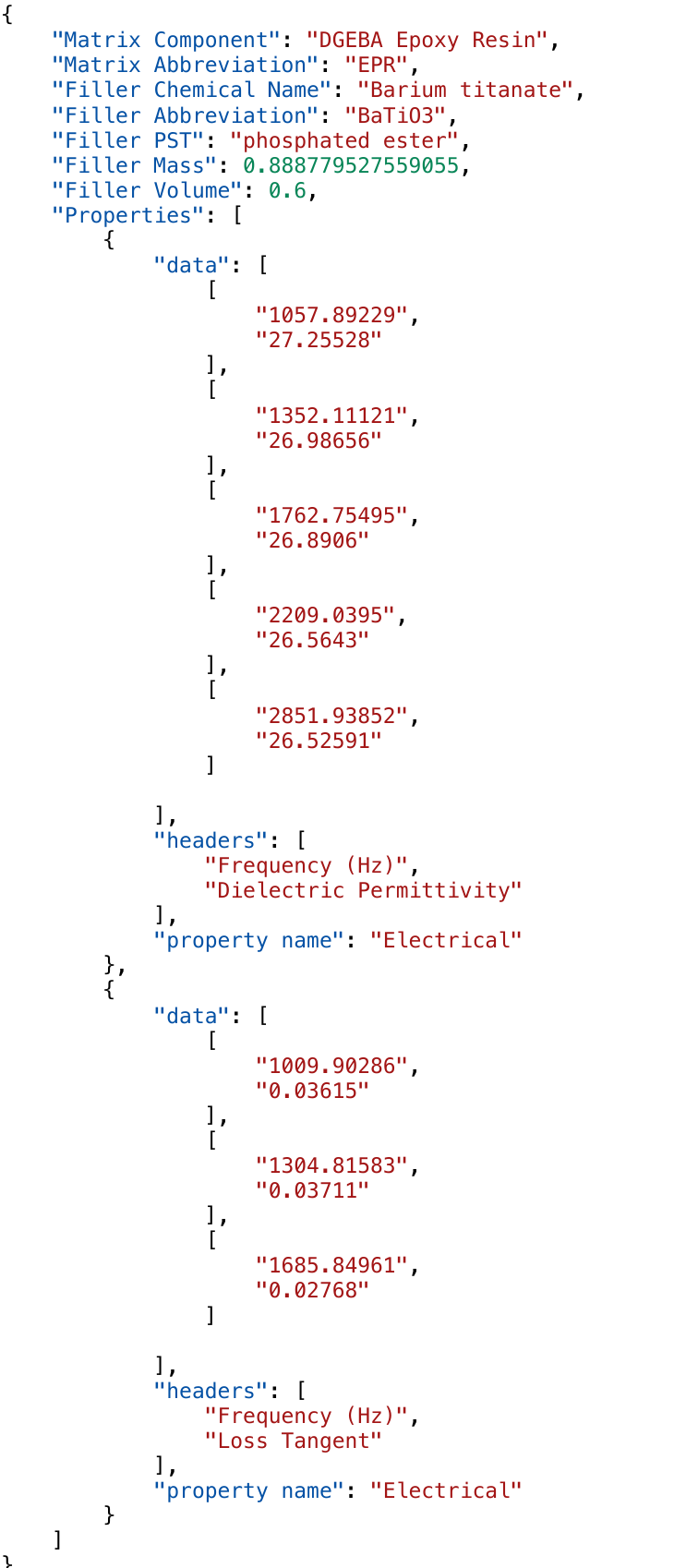}
    \end{subfigure}
    \caption{A figure and its corresponding sample. Note how the data points in the properties are coming from the plot in the image. Also note that the data points in the JSON are shortened to fit on the page; the actual JSON is much larger. Some information in the JSON, like the full name of the filler PST, is not shown in the figure but can be found in the text. See the original article~\citep{DANG2008171}.}

    \label{app:example_pnc_properties}
\end{figure*}
\begin{figure*}[t!]
\subsection{PBD}
    \centering
    \begin{subfigure}[b]{0.45\textwidth}
        \centering
        \includegraphics[width=\textwidth]{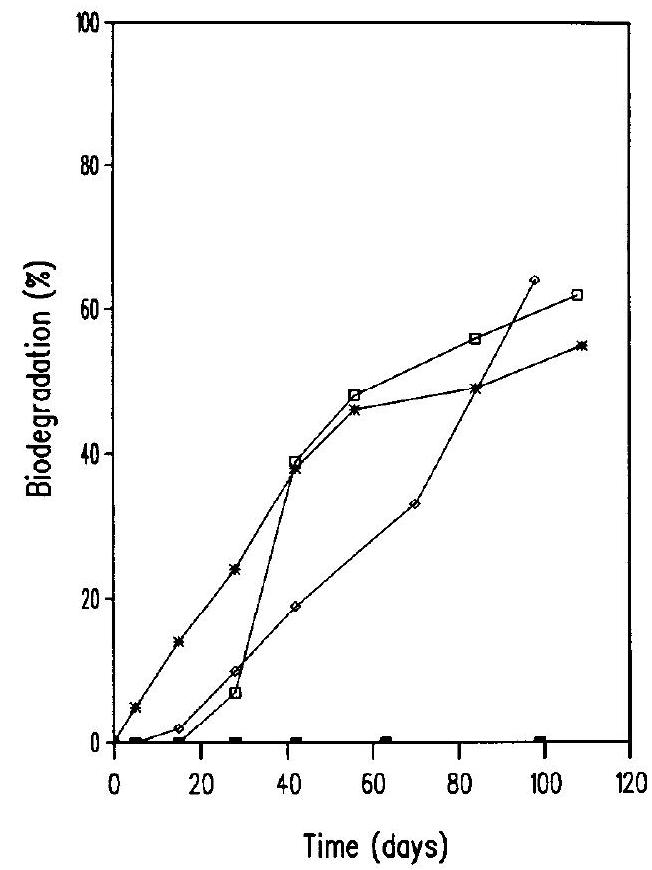}
    \end{subfigure}
    \hfill
    \begin{subfigure}[b]{0.45\textwidth}
        \centering
        \includegraphics[width=\textwidth]{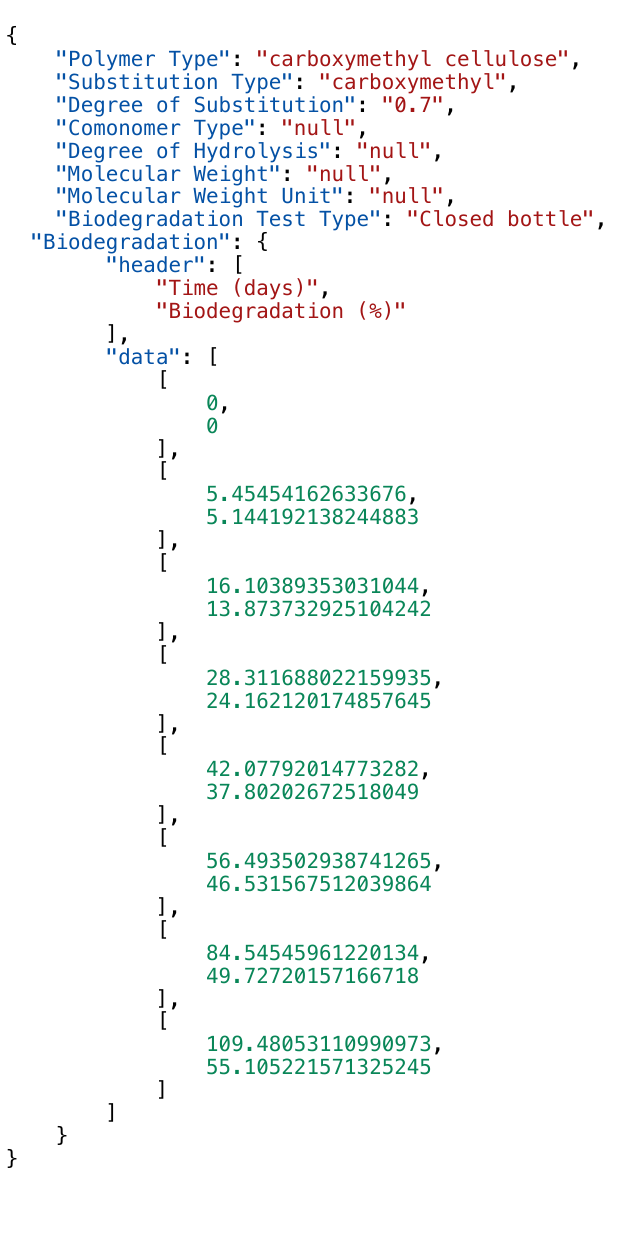}
    \end{subfigure}
    \caption{A figure and its corresponding sample. Note how the data points in the properties are derived from the plot in the image. There are three types of data points in this plot; while these are not explicitly labeled in the image, the figure title specifies which samples each type corresponds to. See the original article~\citep{VANGINKEL1992319}.}

    \label{app:example_pbd_properties}
\end{figure*}

\begin{figure*}[t!]
\section{Annotation Guidelines for PBD Papers}
\centering
\begin{tcolorbox}[colback=gray!10!white, colframe=black!70!white, boxrule=0.4pt, width=\linewidth, arc=2mm, outer arc=2mm]
\ttfamily
\footnotesize
\paragraph{Article Selection and Overview}
\begin{itemize}
    \item The dataset should be composed of high-quality research papers focused on polymer biodegradation, with a preference for those published in reputable journals.
    \item Make sure that the selection includes studies primarily investigating biodegradable materials. Review articles summarizing data from various studies are also acceptable. Ensure access to all selected articles through institutional or open-access resources. 
    \item Organize each paper as a separate folder. Name each folder using a key.
\end{itemize}

\paragraph{Attribute Identification}

\begin{itemize}
    \item Find attributes that are critical in defining material compositions and are consistently reported across articles.
\end{itemize}

\paragraph{Annotation Process for Samples}

\begin{itemize}
    \item For papers containing multiple samples, create a separate JSON file for each sample within the folder corresponding to the paper. Ensure that each JSON file contains only one value per field (e.g., do not combine multiple DS values in a single file).

    \item Use the structure defined in the provided JSON format (see Appendix~\ref{app:json_formats}).
\end{itemize}

\paragraph{Composition and Property Extraction}
\begin{itemize}
    \item Carefully read the text, tables, and figures in each paper to extract information about the composition and properties of the samples. Ensure that common components like polymer type, substitution type, and test conditions are consistently annotated.
    \item For properties appearing as figures (e.g., biodegradation plots), use the PlotDigitizer tool:
    \begin{itemize}
        \item Upload the plot image from the article into PlotDigitizer.
        \item Calibrate the axes by marking reference points (e.g., labeled ticks or values).
        \item Trace the curve to extract numerical data points as (x, y) pairs.
        \item Convert these extracted values into the structured JSON format specified.
    \end{itemize}
\end{itemize}

\paragraph{Validation and Quality Control}
\begin{itemize}
    \item The first annotator should extract the data, while a second annotator reviews the annotations for accuracy. 
    \item If discrepancies arise, the team should discuss and resolve them.
\end{itemize}

\paragraph{Notes on Special Cases}
\begin{itemize}
    \item If images are too complex or data is not clearly labeled, extract a few representative data points to provide a manageable subset for analysis.
\end{itemize}

\paragraph{Final Checks and Updates}
\begin{itemize}
    \item Review annotations periodically to confirm that all attributes align with the defined guidelines.
    \item If new important attributes or inconsistencies are identified, update the guidelines accordingly
\end{itemize}
\end{tcolorbox}
\caption{Annotation guidelines for identifying PBD sample compositions and properties.}
\label{app:annotation_guidelines}
\end{figure*}



    

\begin{figure*}[t!]
\section{Prompts}
\subsection{Text-only Input}
\centering
\begin{tcolorbox}[colback=gray!10!white, colframe=black!70!white, boxrule=0.4pt, width=\linewidth, arc=2mm, outer arc=2mm]
\ttfamily
\footnotesize
\textbf{PROMPT}\vspace{0.2cm}

You extract structured data from scientific articles.\vspace{0.2cm}

Please read the following paragraphs, find all the nano-composite samples, and fill out the given JSON template for each one of those nanocomposite samples. Do not merge samples of different compositions. If an attribute is not mentioned, fill that section with "null". Mass and Volume Composition should be followed by a \%.\vspace{0.3cm}

\textbf{JSON Template:}

\begin{tcolorbox}[colback=white, colframe=black!40!white, boxrule=0.5pt, width=\linewidth, arc=2mm, outer arc=2mm]
\centering
\includegraphics[width=\linewidth]{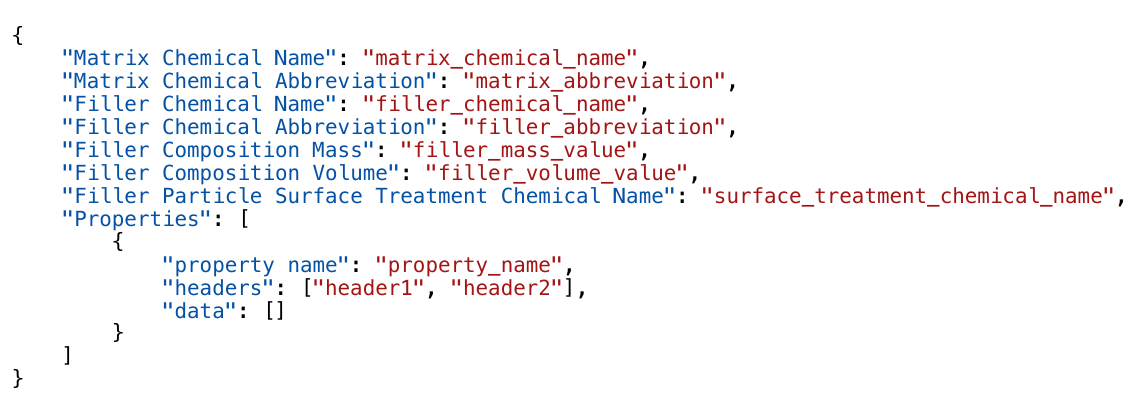}
\end{tcolorbox}

Properties is a list of dictionaries where each dictionary represents a property of the nanocomposite. The property name should be filled out with the name of the property where the choicese are: electrical, mechanical, viscoealstic, thermal, volumetric, rheological.
The headers should be filled out with the x and y labels which are the names of the conditions or the labels of the data (e.g. time, temperature, frequency, strain, conductivity, dielectric strength, etc.). The data should be a list of (x, y) tuples. For example, if the property is 24 MPa at temperature 25°C and 30 MPa at temperature 50°C, the data should be [(25, 24), (50, 30)]. If no data is mentioned, please fill it with null.

\vspace{0.3cm}

\textbf{Article:}

\begin{tcolorbox}[colback=white, colframe=black!40!white, boxrule=0.5pt, width=\linewidth, arc=2mm, outer arc=2mm]
\centering
\includegraphics[width=\linewidth]{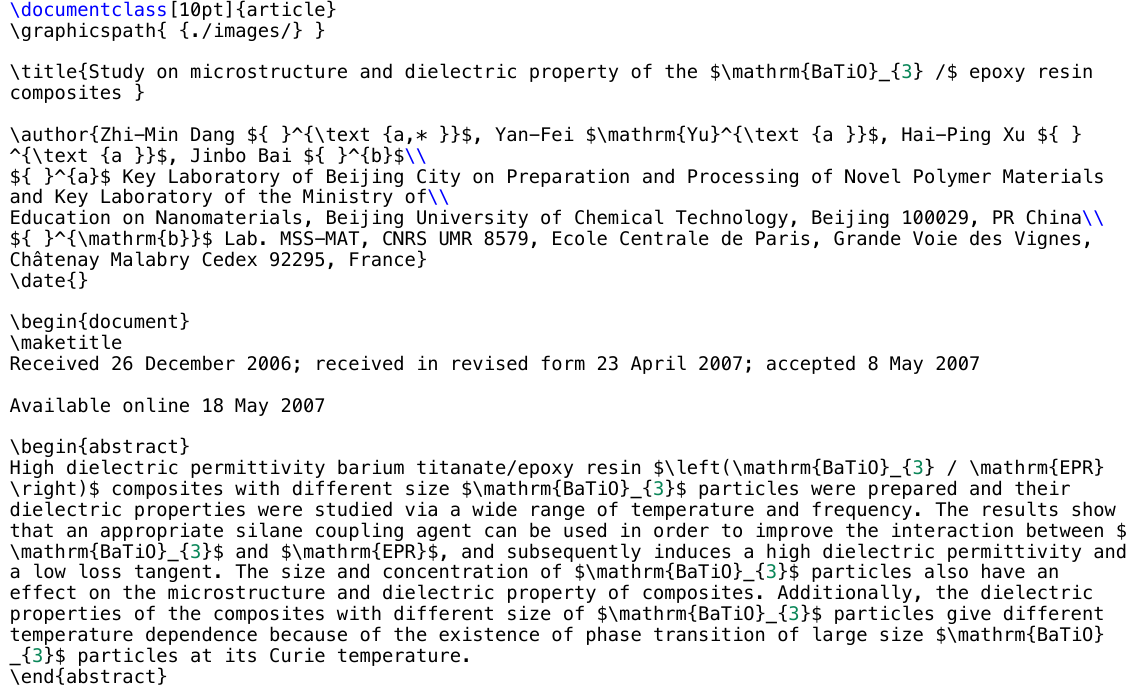}
\end{tcolorbox}

\end{tcolorbox}
\caption{Sample prompt to GPT-4o for extracting nanocomposite samples. The provided article has been truncated due to space constraints. The input is given without including the figures parsed by DePlot and without providing any images—only textual input was given to the model.}
\label{app:nanocomposite_extraction}
\end{figure*}

\begin{figure*}[t!]
\subsection{LLM Output}
\centering
\begin{tcolorbox}[colback=gray!10!white, colframe=black!70!white, boxrule=0.4pt, width=\linewidth, arc=2mm, outer arc=2mm]
\ttfamily
\footnotesize
\textbf{GPT-4o RESPONSE}\vspace{0.2cm}

Based on the document, one nanocomposite sample was identified:

\begin{tcolorbox}[colback=white, colframe=black!40!white, boxrule=0.5pt, width=\linewidth, arc=2mm, outer arc=2mm]
\centering
\includegraphics[width=\linewidth]{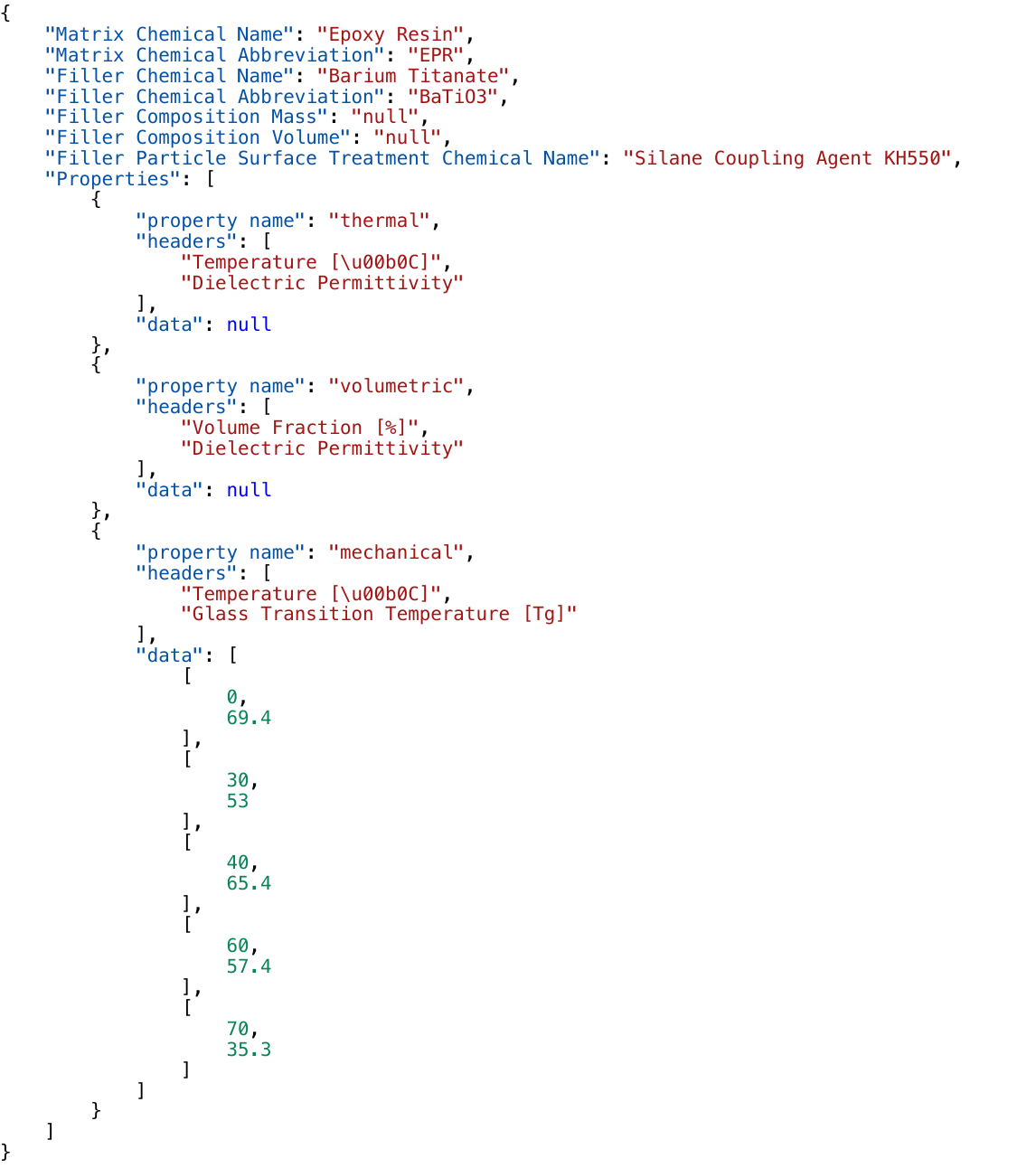}
\end{tcolorbox}

Note: The document provides detailed thermal and mechanical properties, but specific mass and volume compositions are not mentioned, hence "null" in the corresponding fields.
\end{tcolorbox}
\caption{GPT-4o response to the nanocomposite data extraction task described in Figure~\ref{app:nanocomposite_extraction}.}
\label{app:gpt4_response}
\end{figure*}

\begin{figure*}[t!]
\subsection{Text+Image Input}
\centering
\begin{tcolorbox}[colback=gray!10!white, colframe=black!70!white, boxrule=0.4pt, width=\linewidth, arc=2mm, outer arc=2mm]
\ttfamily
\footnotesize
\textbf{PROMPT}\vspace{0.2cm}

Given the image and the following polymer nanocomposite sample compositions, first identify which sample composition is present in the image, then extract its properties.\vspace{0.3cm}

\textbf{Sample Compositions:}

\begin{tcolorbox}[colback=white, colframe=black!40!white, boxrule=0.5pt, width=\linewidth, arc=2mm, outer arc=2mm]
\centering
\includegraphics[width=\linewidth]{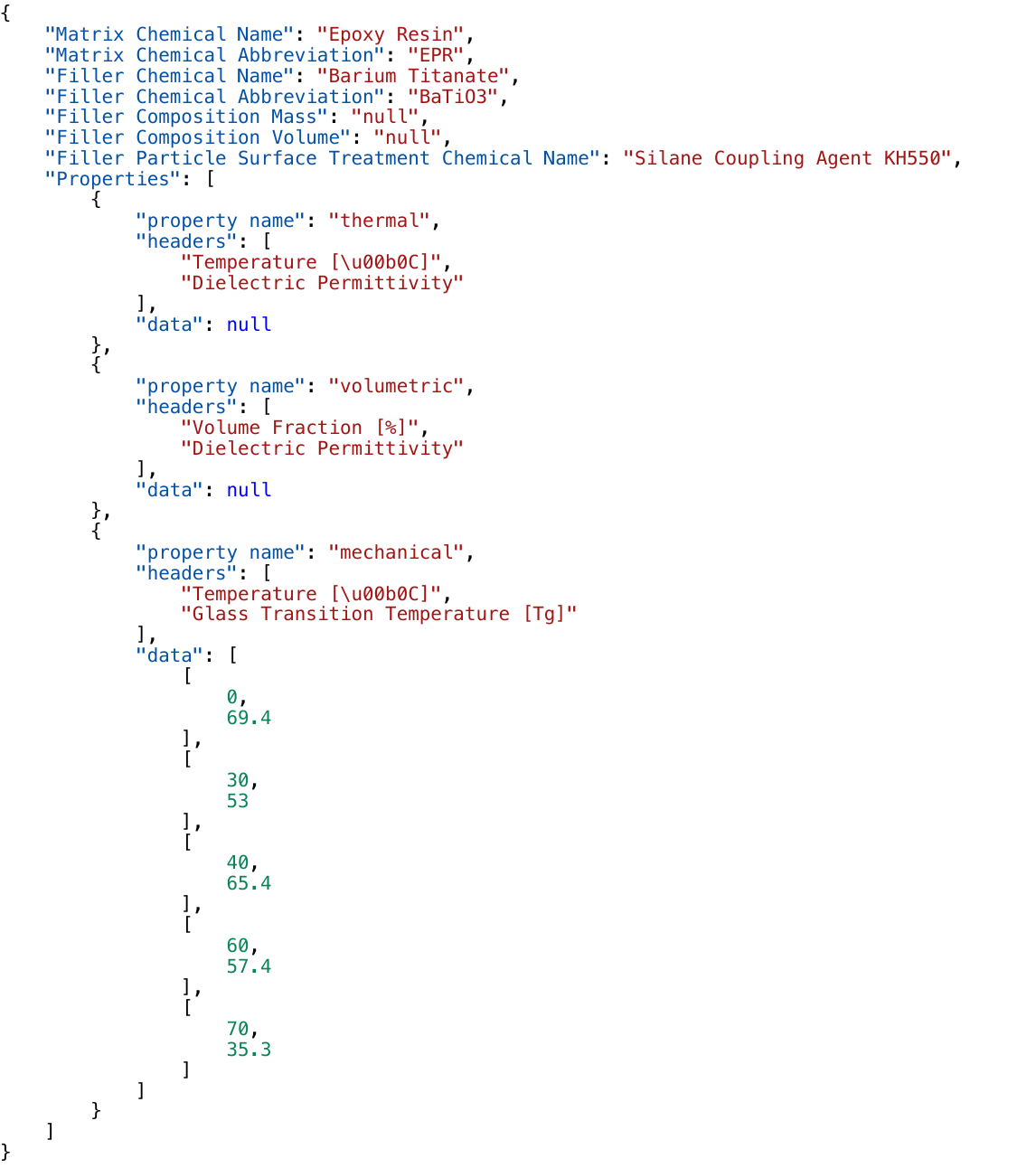}
\end{tcolorbox}

For those sample compositions that are present in the image, extract the information about the property.

Properties is a list of dictionaries where each dictionary represents a property of the nanocomposite. The property name should be filled out with the name of the property where the choicese are: electrical, mechanical, viscoealstic, thermal, volumetric, rheological.
The headers should be filled out with the x and y labels which are the names of the conditions or the labels of the data (e.g. time, temperature, frequency, strain, conductivity, dielectric strength, etc.). The data should be a list of (x, y) tuples. For example, if the property is 24 MPa at temperature 25°C and 30 MPa at temperature 50°C, the data should be [(25, 24), (50, 30)]. If no data is mentioned, please fill it with null.

Expand the sample composition JSONs to include the property information and return all the expanded JSONs.
\end{tcolorbox}
\caption{Sample prompt to GPT-4o for identifying polymer nanocomposite sample compositions from an image and extracting their properties.}
\label{app:nanocomposite_extraction_image}
\end{figure*}

\begin{figure*}[t!]
\subsection{VLM Output}
\centering
\begin{tcolorbox}[colback=gray!10!white, colframe=black!70!white, boxrule=0.4pt, width=\linewidth, arc=2mm, outer arc=2mm]
\ttfamily
\footnotesize
\textbf{GPT-4o RESPONSE}\vspace{0.2cm}

The image shows dielectric permittivity and loss tangent as a function of temperature for two different loadings of BaTiO3 in an epoxy resin matrix. However, the sample composition provided does not specify the particle size or volume fraction, so we cannot directly match it to the image. 

Given the sample composition provided, we can expand the JSON to include the property information as follows:

\begin{tcolorbox}[colback=white, colframe=black!40!white, boxrule=0.5pt, width=\linewidth, arc=2mm, outer arc=2mm]
\centering
\includegraphics[width=\linewidth]{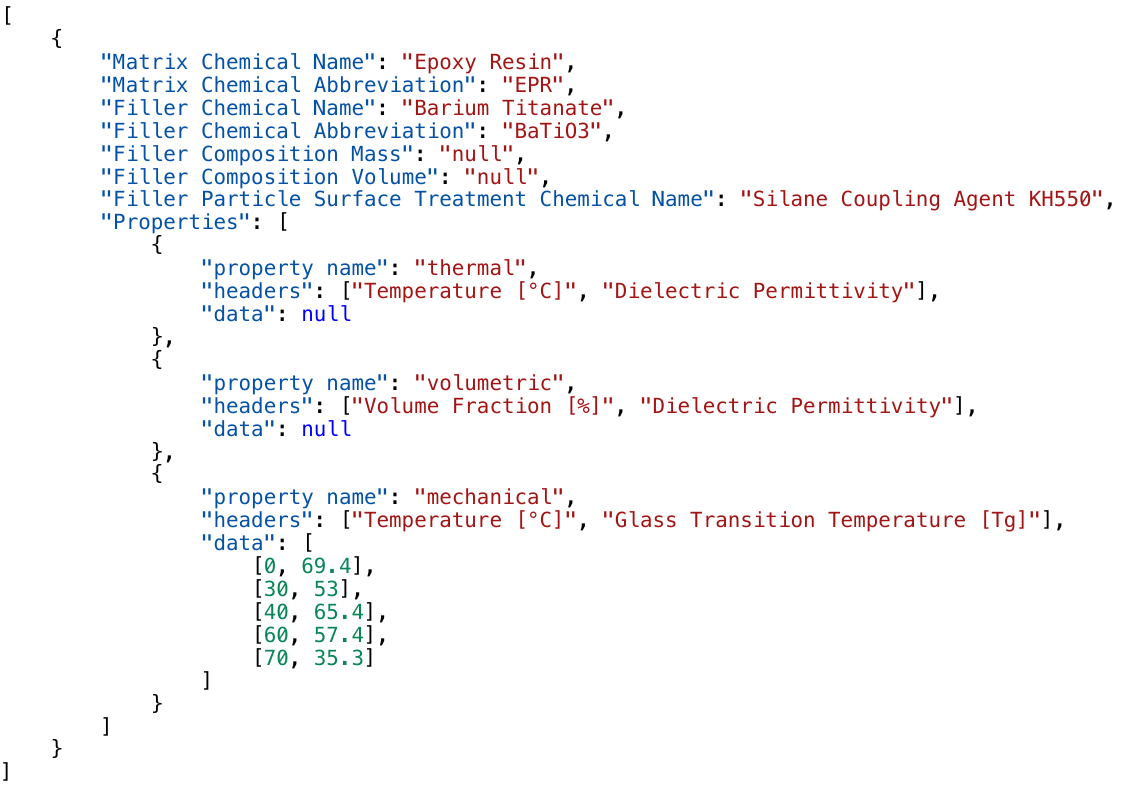}
\end{tcolorbox}

\end{tcolorbox}
\caption{GPT-4o response to the nanocomposite data extraction task described in Figure~\ref{app:nanocomposite_extraction_image}, given one of the images from the paper. Note that if there are \(n\) images in the document, there will be \(n\) separate responses for each image, which will later be merged together.}
\label{app:gpt4_response_image}
\end{figure*}

\begin{figure*}[t!]
\section{Human Evaluation Guideline}
\centering
\begin{tcolorbox}[colback=gray!10!white, colframe=black!70!white, boxrule=0.4pt, width=\linewidth, arc=2mm, outer arc=2mm]
\ttfamily
\footnotesize
\textbf{Instructions} 

For each plot pair, please provide a score from 1 (poor) to 5 (excellent) for the following two aspects:

\begin{itemize}
    \item Axis Label Accuracy:
    \begin{itemize}
        \item Check if the labels on the x-axis and y-axis of the predicted plot match those of the ground-truth plot.
        \item 1: Labels are entirely incorrect or missing.
        \item 2: Labels are mostly incorrect, with one or two minor matches.
        \item 3: Labels are partially correct (e.g., one axis matches, the other is incorrect).
        \item 4: Labels are mostly correct, with only minor errors (e.g., small formatting differences).
        \item 5: Labels are completely correct and match perfectly with the ground-truth plot.
    \end{itemize}
    \item Curve Trend Consistency:
    \begin{itemize}
        \item 1: The trend is completely different from the ground truth.
        \item 2: The trend shows some alignment but is mostly inconsistent.
        \item 3: The trend matches in some sections.
        \item 4: The trend is mostly consistent, with only minor deviations.
        \item 5: The trend is fully consistent and matches the ground truth perfectly.
    \end{itemize}
\end{itemize}

\includegraphics[scale=0.5]{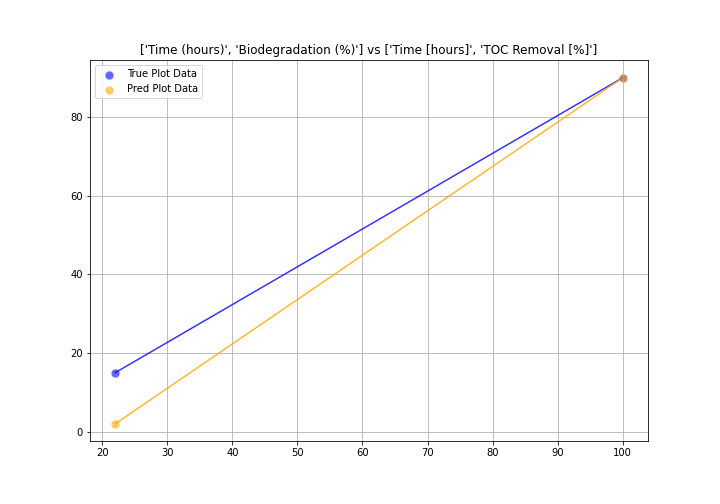}
\end{tcolorbox}
\caption{Scoring guidelines for human evaluation.}
\label{app:human_eval_guideline}
\end{figure*}

\end{document}